\documentclass[conference]{IEEEtran}     %

\IEEEoverridecommandlockouts    %

\usepackage{amsmath} %
\usepackage[pdftex]{graphicx}
\usepackage{tabularx}
\graphicspath{{figures/}}
\usepackage{hyperref}
\hypersetup{
    colorlinks=true,
    linkcolor=black,
    citecolor=black,
    filecolor=black,
    urlcolor=black,
}
\usepackage[T1]{fontenc}
\usepackage[utf8]{inputenc}
\usepackage{csquotes}
\usepackage[english]{babel}
\usepackage[export]{adjustbox}
\usepackage{caption}
\usepackage{balance}
\usepackage{subcaption}
\usepackage[colorinlistoftodos]{todonotes}
\usepackage{lipsum}
\usepackage{siunitx}
\usepackage{tikz, pgfplots, pgfplotstable}
\usepackage{textcomp}
\usetikzlibrary{plotmarks}
\usepackage{enumerate}
\usepackage{placeins}%
\usepackage{lipsum}
\usepackage{amssymb}
\usepackage{multirow}

\definecolor{darkraspberry}{rgb}{0.53, 0.15, 0.34}
\definecolor{electricyellow}{rgb}{1.0, 1.0, 0.0}
\definecolor{gold(web)(golden)}{rgb}{1.0, 0.84, 0.0}

\usepackage[style=ieee,
            doi=false,
            url=false,
            mincitenames=1,
            maxcitenames=1,
            minbibnames=6,
            maxbibnames=6,
            backend=biber]{biblatex}  %
\AtBeginBibliography{\small}

\title{Vision-based Lifting of 2D Object Detections for Automated Driving\\
}

\author{\IEEEauthorblockN{Hendrik Königshof}
\IEEEauthorblockA{\textit{Intel. Systems and Prod. Engineering} \\
\textit{FZI Research Center for Inf. Technology}\\
Karlsruhe, Germany \\
koenigshof@fzi.de}
\and
\IEEEauthorblockN{Kun Li}
\IEEEauthorblockA{\textit{Intel. Systems and Prod. Engineering} \\
\textit{FZI Research Center for Inf. Technology}\\
Karlsruhe, Germany \\
li2@fzi.de}
\and
\IEEEauthorblockN{Christoph Stiller}
\IEEEauthorblockA{\textit{Institute of Meas. and Control Systems} \\
\textit{Karlsruhe Institute of Technology (KIT)}\\
Karlsruhe, Germany \\
stiller@kit.edu}
}

\usepackage{textcomp}
\usepackage[absolute,showboxes]{textpos}
\usepackage{setspace}

\TPMargin{5pt}

\newcommand{\acceptancenotice}[2]{  %
    \begin{textblock}{0.82}(0.09,0.935)
        \setstretch{0.65} %
        \noindent{\footnotesize{\copyright #1 IEEE.
        Personal use of this material is permitted.
        Permission from IEEE must be obtained for all other uses, in any current or future media,
        including reprinting/republishing this material for advertising or promotional purposes,
        creating new collective works, for resale or redistribution to servers or lists,
        or reuse of any copyrighted component of this work in other works.
        DOI: \url{#2}
        \noindent
        }}
    \end{textblock}
}

\begin{document}
\maketitle

\acceptancenotice{2020}{https://doi.org/10.23919/FUSION45008.2020.9190325}

\thispagestyle{empty}
\pagestyle{empty}

\begin{abstract}

Image-based 3D object detection is an inevitable part of autonomous driving because cheap onboard cameras are already available in most modern cars.
Because of the accurate depth information, currently most state-of-the-art 3D object detectors heavily rely on LiDAR data. 
In this paper, we propose a pipeline which lifts the results of existing vision-based 2D algorithms to 3D detections using only cameras as a cost-effective alternative to LiDAR. 
In contrast to existing approaches, we focus not only on cars but on all types of road users. To the best of our knowledge, we are the first using a 2D CNN to process the point cloud for each 2D detection to keep the computational effort as low as possible.
Our evaluation on the challenging KITTI 3D object detection benchmark shows results comparable to state-of-the-art image-based approaches while having a runtime of only a third.

\end{abstract}
 

\section{Introduction}
\label{sec:introduction}

With just one glance at an image, humans are able to determine which objects are in the image and where they are.
The human visual system is fast and accurate, allowing us to perform complex tasks like driving a car relatively easily.
In order to enable autonomous driving, 3D detection algorithms with similar reliability and robustness are a fundamental requirement.
Compared to other types of sensors, camera-based approaches offer the most cost-effective solution for such a task.
State-of-the-art algorithms already achieve up to 95\% average precision for car detections in 2D images, nevertheless, there are still only a few approaches using only cameras to expand this information from 2D to 3D.

In this work, we present a vision-based pipeline to lift 2D detections to 3D using semantic and spatial information.
Our main contributions are the following:

1) We present the first approach to process mono or stereo point clouds using a 2D Convolutional Neural Network (CNN) architecture and reaching results comparable with state-of-the-art approaches;

2) We introduce a lifting pipeline that can be used with any 2D detection network providing reliable results due to its shift-invariance even for poor 2D detections;

The paper is organized as follows: In the next section we summarize popular LiDAR and camera-based object detectors and motivate our approach. In Section~\ref{sec:system_overview} an overview of our method is given. Subsequently, the lifting process to attain the 3D bounding boxes, is discussed in more detail in Section~\ref{sec:lifting_2d_detection_to_3d}. Evaluation on real-world data is presented in Section~\ref{sec:results_and_evaluation} before the paper is concluded by a summary and an outlook.

\begin{figure}[t]
	\begin{subfigure}{0.966\columnwidth}
		\setlength{\abovecaptionskip}{6pt plus 2pt minus 2pt}
		\setlength{\belowcaptionskip}{7pt plus 2pt minus 2pt}
		\includegraphics[width=\linewidth]{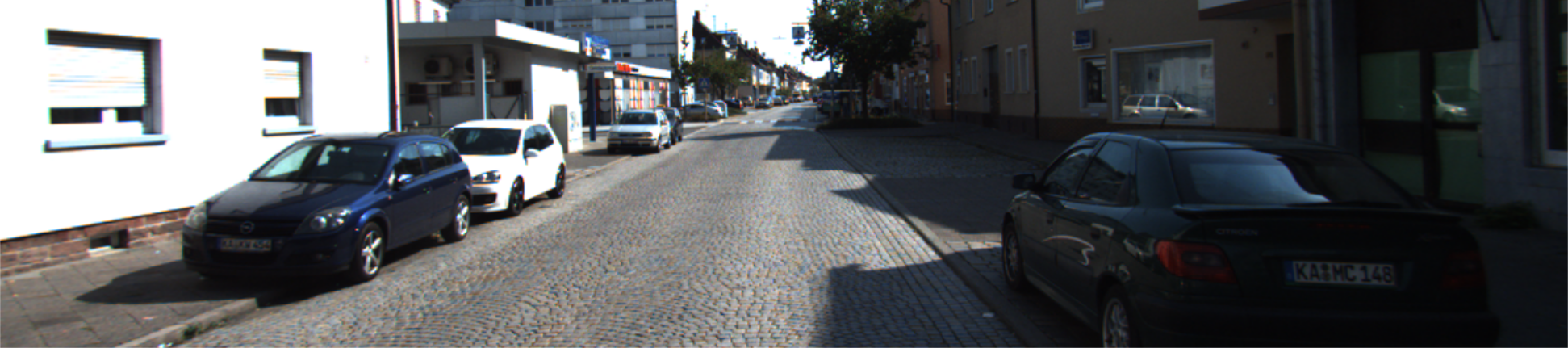}
		\caption{Input Image}
		\label{fig:liftingProcess_a}
	\end{subfigure}
	\begin{subfigure}{0.966\columnwidth}
		\setlength{\abovecaptionskip}{6pt plus 2pt minus 2pt}
		\setlength{\belowcaptionskip}{7pt plus 2pt minus 2pt}
		\includegraphics[width=\linewidth]{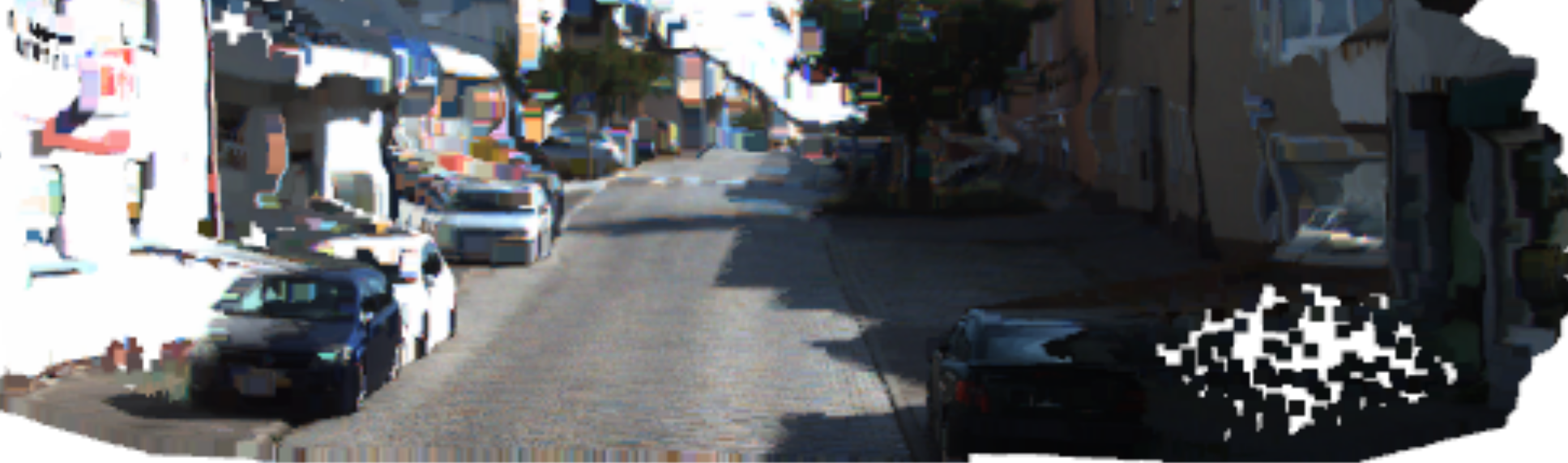}
		\caption{Point Cloud}
		\label{fig:liftingProcess_b}
	\end{subfigure}
	\begin{subfigure}{0.966\columnwidth}
		\setlength{\abovecaptionskip}{6pt plus 2pt minus 2pt}
		\setlength{\belowcaptionskip}{7pt plus 2pt minus 2pt}
		\includegraphics[width=\linewidth]{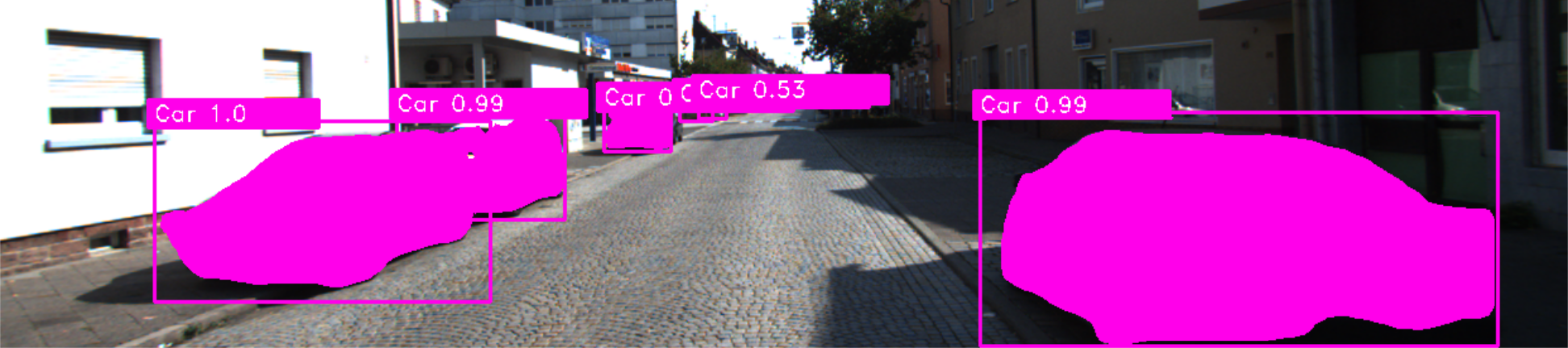}
		\caption{2D Detections with Instance Masks}
		\label{fig:liftingProcess_c}
	\end{subfigure}
	\begin{subfigure}{\columnwidth}
		\setlength{\abovecaptionskip}{-12pt plus 2pt minus 2pt}
		\setlength{\belowcaptionskip}{7pt plus 2pt minus 2pt}
		\includegraphics[width=\linewidth]{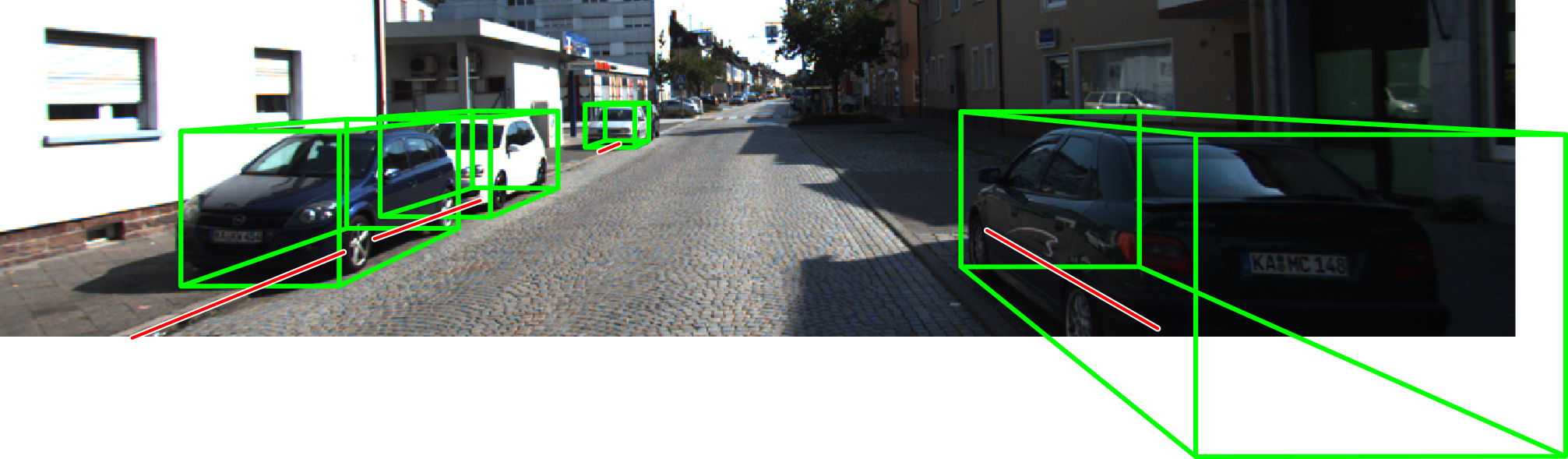}
		\caption{3D Bounding Boxes}
		\label{fig:liftingProcess_d}
	\end{subfigure}
	\caption{Given an image (\subref{fig:liftingProcess_a}), we obtain the organized point cloud (\subref{fig:liftingProcess_b}). After estimating 2D bounding boxes, semantic segmentation is performed for each 2D detection (\subref{fig:liftingProcess_c}). Both semantic and spatial information of the 2D detections is further processed by a CNN to attain the parameters for the corresponding 3D bounding boxes (\subref{fig:liftingProcess_d}). Red lines indicate object orientations.}
	\label{fig:liftingProcess}
\end{figure}

\begin{figure*}[t]
\centering
\hspace*{-1cm}
\fontsize{8}{10}\selectfont
\def\svgwidth{\textwidth}
\begingroup%
  \makeatletter%
  \providecommand\color[2][]{%
    \errmessage{(Inkscape) Color is used for the text in Inkscape, but the package 'color.sty' is not loaded}%
    \renewcommand\color[2][]{}%
  }%
  \providecommand\transparent[1]{%
    \errmessage{(Inkscape) Transparency is used (non-zero) for the text in Inkscape, but the package 'transparent.sty' is not loaded}%
    \renewcommand\transparent[1]{}%
  }%
  \providecommand\rotatebox[2]{#2}%
  \newcommand*\fsize{\dimexpr\f@size pt\relax}%
  \newcommand*\lineheight[1]{\fontsize{\fsize}{#1\fsize}\selectfont}%
  \ifx\svgwidth\undefined%
    \setlength{\unitlength}{542.09976209bp}%
    \ifx\svgscale\undefined%
      \relax%
    \else%
      \setlength{\unitlength}{\unitlength * \real{\svgscale}}%
    \fi%
  \else%
    \setlength{\unitlength}{\svgwidth}%
  \fi%
  \global\let\svgwidth\undefined%
  \global\let\svgscale\undefined%
  \makeatother%
  \begin{picture}(1,0.41428151)%
    \lineheight{1}%
    \setlength\tabcolsep{0pt}%
    \put(0,0){\includegraphics[width=\unitlength,page=1]{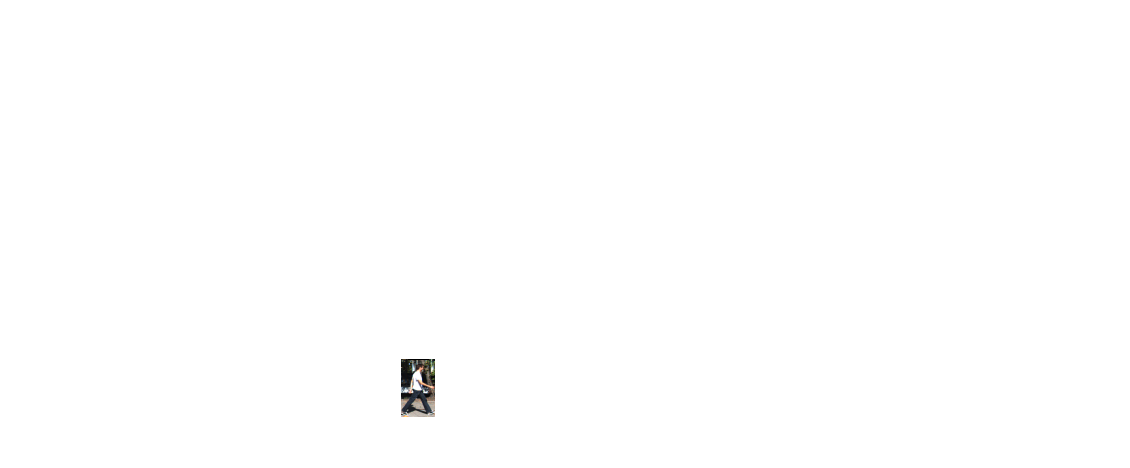}}%
    \put(0,0){\includegraphics[width=\unitlength,page=2]{pipeline_detailed_000000_final3.pdf}}%
    \put(0,0){\includegraphics[width=\unitlength,page=3]{pipeline_detailed_000000_final3.pdf}}%
    \put(0.31904376,0.03099961){\color[rgb]{0,0,0}\makebox(0,0)[lt]{\lineheight{1.25000012}\smash{\begin{tabular}[t]{l}Regions of Interest\end{tabular}}}}%
    \put(0,0){\includegraphics[width=\unitlength,page=4]{pipeline_detailed_000000_final3.pdf}}%
    \put(0.07620926,0.15036321){\color[rgb]{1,0,0}\makebox(0,0)[lt]{\lineheight{1.25}\smash{\begin{tabular}[t]{l}Mono Input Image\\\end{tabular}}}}%
    \put(0.33679592,0.21396979){\color[rgb]{0,0,1}\makebox(0,0)[lt]{\lineheight{1.25}\smash{\begin{tabular}[t]{l}Point Cloud\\ Generation\end{tabular}}}}%
    \put(0.31211127,0.00281171){\color[rgb]{0,0,1}\makebox(0,0)[lt]{\lineheight{1.25}\smash{\begin{tabular}[t]{l}2D Object Detection\end{tabular}}}}%
    \put(0.52690637,0.02311925){\color[rgb]{0,0,1}\makebox(0,0)[lt]{\lineheight{1.25}\smash{\begin{tabular}[t]{l}Semantic\\Segmentation\end{tabular}}}}%
    \put(0.07414809,0.31539643){\color[rgb]{0.5,0.5,0.5}\makebox(0,0)[lt]{\lineheight{1.25}\smash{\begin{tabular}[t]{l}Stereo Input Image\\\end{tabular}}}}%
    \put(0.28867266,0.32857263){\color[rgb]{0,0,0}\makebox(0,0)[lt]{\lineheight{1.25}\smash{\begin{tabular}[t]{l}Estimated Disparity(Depth) \end{tabular}}}}%
    \put(0.34185362,0.23828297){\color[rgb]{0,0,0}\makebox(0,0)[lt]{\lineheight{1.25}\smash{\begin{tabular}[t]{l}Point Cloud\end{tabular}}}}%
    \put(0,0){\includegraphics[width=\unitlength,page=5]{pipeline_detailed_000000_final3.pdf}}%
    \put(0.31433121,0.09805019){\color[rgb]{0,0,0}\makebox(0,0)[lt]{\lineheight{1.25}\smash{\begin{tabular}[t]{l}2D Bounding Boxes\end{tabular}}}}%
    \put(0,0){\includegraphics[width=\unitlength,page=6]{pipeline_detailed_000000_final3.pdf}}%
    \put(0.65168866,0.16020091){\color[rgb]{0,0,0}\makebox(0,0)[lt]{\lineheight{1.25}\smash{\begin{tabular}[t]{l}Concatenated data\end{tabular}}}}%
    \put(0.66604285,0.07625361){\color[rgb]{0,0,0}\makebox(0,0)[lt]{\lineheight{1.25}\smash{\begin{tabular}[t]{l}Resized to 64x64\end{tabular}}}}%
    \put(0,0){\includegraphics[width=\unitlength,page=7]{pipeline_detailed_000000_final3.pdf}}%
    \put(0.81195023,0.18965729){\color[rgb]{0,0,0}\makebox(0,0)[lt]{\lineheight{1.25}\smash{\begin{tabular}[t]{l}Modified \\ResNet50\end{tabular}}}}%
    \put(0,0){\includegraphics[width=\unitlength,page=8]{pipeline_detailed_000000_final3.pdf}}%
    \put(0.71161532,0.272684){\color[rgb]{0,0,0}\makebox(0,0)[lt]{\lineheight{1.25}\smash{\begin{tabular}[t]{l}Lifted 3D Bounding Box\end{tabular}}}}%
    \put(0.69627199,0.05202535){\color[rgb]{0,1,0}\makebox(0,0)[lt]{\lineheight{1.25}\smash{\begin{tabular}[t]{l}Lifting 2D Detection to 3D\end{tabular}}}}%
    \put(0,0){\includegraphics[width=\unitlength,page=9]{pipeline_detailed_000000_final3.pdf}}%
    \put(0.76730786,0.11008352){\color[rgb]{0,0,0}\makebox(0,0)[lt]{\begin{minipage}{0.08301055\unitlength}\raggedright \end{minipage}}}%
    \put(0.0004485,0.50141898){\color[rgb]{0,0,0}\makebox(0,0)[lt]{\begin{minipage}{1.0208321\unitlength}\raggedright \end{minipage}}}%
    \put(0.86316527,0.09624842){\color[rgb]{0,0,0}\makebox(0,0)[lt]{\begin{minipage}{0.17392688\unitlength}\raggedright \end{minipage}}}%
    \put(0.16251671,0.33737432){\color[rgb]{0,0,0}\makebox(0,0)[lt]{\begin{minipage}{1.07320789\unitlength}\raggedright \end{minipage}}}%
    \put(-0.03018635,0.5310656){\color[rgb]{0,0,0}\makebox(0,0)[lt]{\begin{minipage}{0.26681964\unitlength}\raggedright \end{minipage}}}%
  \end{picture}%
\endgroup%
 \caption{Pipeline to lift 2D detections to 3D.
Our pipeline is suitable for both \textcolor{red}{mono} and \textcolor{gray}{stereo} vision.
The pipeline consist of three basic blocks (\textcolor{blue}{2D object detection, point cloud generation, semantic segmentation}) and a core block (\textcolor{green}{lifting 2D detection to 3D}).
Depth estimation and 2D object detection can be performed in parallel to save runtime.
In our work, the semantic segmentation was done for every ROI, which is obtained from the 2D detection.
The lifting model takes spatial and semantic information as input and lifts the 2D detection to 3D.
}
\label{fig:pipeline}
\end{figure*} %
\section{Related Work}
\label{sec:related_work}
Recently, significant progress has been made in 3D object detection.
Due to the high depth estimation accuracy of LiDAR, many techniques have been proposed, which apply deep learning on point clouds.
Frustum PointNet \cite{8578200}, for example, attains a region of interest (ROI) proposal from a 2D image.
By extracting the ROI from the point cloud, a frustum point cloud is gained, on which the PointNet \cite{8099499} is applied for further object detection.
MV3D \cite{8100174} encodes the sparse 3D point cloud with a compact multi-view representation.
VoxelNet \cite{8578570} divides a point cloud into equally spaced 3D voxels and transforms a group of points within each voxel into a unified feature representation.
UberATG-ContFuse \cite{Liang_2018_ECCV} exploits continuous convolutions to fuse image and LiDAR feature maps at different levels of resolution.
All these approaches are based on 3D point clouds obtained by LiDAR.

Since LiDAR is very expensive compared to cameras, efforts have been made to find a proper distance measuring method based on cameras to replace LiDAR.
Methods have been proposed for monocular \cite{8578312}\cite{8100182}\cite{Yin_2019_ICCV} and stereo vision \cite{7780807}\cite{8578665}.
Among those, the PSMNet \cite{8578665} ranked first in the KITTI \cite{6248074} 2012 and 2015 leaderboards before March 2018 and VNLNet \cite{Yin_2019_ICCV} is one of the leading algorithms for mono vision on KITTI \cite{6248074}.

The improvement of monocular and stereo depth estimation enabled 3D object detection by only using cameras.
In \cite{wang2018pseudolidar}, the authors use DORN \cite{8578312} and PSMNet \cite{8578665} to generate pseudo point clouds, which are further used as input for Frustum PointNet \cite{8578200} and AVOD \cite{8594049} for 3D object detection.
In this approach, the backbone for object detection is the same as when using LiDAR point clouds.

Other approaches try to lift 2D detections to 3D.
The approach in \cite{weng2019monocular} uses the instance segmentation and pseudo point cloud of the image to lift the 2D detection to 3D.
ROI-10D \cite{Manhardt_2019_CVPR} formulates a novel loss function to minimize the pointwise distances of the corresponding points of the bounding boxes. 
MonoFENet \cite{8897727} enhances features of the 2D and 3D stream with the estimated disparity and utilize them for 3D localization with the proposed PointFE network.  

Nevertheless, all the mentioned lifting procedures either process the generated point cloud using PointNet-based networks \cite{wang2018pseudolidar}\cite{weng2019monocular} or fuse the depth (disparity) maps into feature maps for further processing \cite{Manhardt_2019_CVPR}.
Through depth estimation using deep learning methods like \cite{8578312}\cite{8100182}\cite{Yin_2019_ICCV} we obtain a depth map with the same image format as the input image.
Given the setup of the cameras, the image pixels can be transformed into a point cloud as suggested in \cite{wang2018pseudolidar}. 
This indicates that the points can be organized in an image format, which makes it possible to process the points with a CNN as in image processing.
With methods like Faster RCNN \cite{7485869}, Single-Stage-Detector \cite{Liu2016SSDSS} and Yolo \cite{7780460}, promising 2D detection results can be achieved.
Moreover, through semantic segmentation, the detected objects can be labeled pixel by pixel.

This inspires us to propose a novel method to lift 2D detections to 3D, in which we concatenate the point clouds of the 2D object detections and their corresponding semantic segmentation information and process them with a CNN to estimate the corresponding 3D bounding boxes.  %
\section{System Overview}
\label{sec:system_overview}

The pipeline to lift a 2D detection to 3D involves three basic tasks which have been well tackled by other models: \textbf{i)} 2D object detection; \textbf{ii)} semantic segmentation; \textbf{iii)} depth estimation.

\paragraph*{\textbf{2D Object Detection}} Current progress in 2D object detection research provides promising 2D bounding boxes, which can be used as \textit{Region of Interest} (ROI).
Since we evaluate our approach on the KITTI Benchmark \cite{6248074}, we utilize the 2D detection results from \cite{salscheider2019simultaneous}. Their CNN was trained and tested on the KITTI object detection dataset and achieves an average precision of around 70\% for cars and pedestrians. It does not reach state-of-the-art results, but achieves good accuracy while being real-time capable.

\paragraph*{\textbf{Semantic Segmentation}} The semantic segmentation for each ROI is performed by a PSPNet101 model \cite{8100143}, which was pretrained on the PASCAL VOC2012 dataset \cite{pascal-voc-2012} for class segmentation.

\paragraph*{\textbf{Depth Estimation}} We use the PSMNet \cite{8578665} for stereo vision and VNLNet \cite{Yin_2019_ICCV} for mono vision to estimate the depth.
In our evaluation, we compare the 3D object detection results where we accordingly use these two approaches to generate the point clouds. %
\section{Lifting 2D Detection to 3D}
\label{sec:lifting_2d_detection_to_3d}
The core task of our pipeline is the lifting process, where we attain 3D bounding boxes exploiting the given 2D bounding boxes.
As explained in Section~\ref{sec:system_overview}, after depth estimation we attain a point cloud that maintains the same image format as each input image.
Utilizing the 2D detection results, we crop the ROI for semantic segmentation, so that we can distinguish the pixels of detected objects from the background.
In parallel, we attain the point cloud for the ROIs with the same size, similar to the method in Frustum PointNet \cite{8578200}.
Then the semantic information and the spatial coordinates of the ROIs are concatenated and possess the size $H\times W\times (C+3)$, where $H$ and $W$ are the height and width of each ROI and $C$ is the total class number from semantic segmentation.
The additional three channels represent the coordinates $x$, $y$ and $z$ of each pixel.

Since the concatenated crops inherit the image format, we reshape them to the size $64\times 64$ and further process them using ResNet50 \cite{7780459}, followed by a multilayer perceptron (MLP).
Details of our pipeline are shown in Figure~\ref{fig:pipeline}.

\paragraph*{\textbf{Class Prior}} As we obtain the class label through the 2D detection, we use the class prior to form an one-hot vector as an additional input for our lifting model.

\paragraph*{\textbf{Localization Estimation}} We use the position of the central pixel $\boldsymbol{p_{\text{m}}}=(x_{\text{m}},y_{\text{m}},z_{\text{m}})^{\text{T}}$ as the prior and estimate the deviation of the position $\Delta \textbf{p}=\textbf{p}_{\text{g}} - \textbf{p}_{\text{m}}$, where $\textbf{p}_{\text{g}}$ represent the ground truth of the location.

\paragraph*{\textbf{Dimension Estimation}} Since the class label is attained through the 2D detection, we utilize the prior information of the object dimension $\boldsymbol{d}_{\text{prior,k}} = (h_{\text{prior,k}},w_{\text{prior,k}},l_{\text{prior,k}})^\text{T}$ for each certain class $k$, similarly to \cite{8917330}. Our model estimates the dimension deviation $\Delta \boldsymbol{d} = \boldsymbol{d}_{\text{g,k}} - \boldsymbol{d}_{\text{prior,k}}$, where $\boldsymbol{d}_{\text{g,k}}$ represents the ground truth of the dimension for the class $k$.

\paragraph*{\textbf{Orientation Estimation}} We encode the orientation $\theta$ as in \cite{mousavian20173d}.
We transform the $\theta$ into a correction angle $\theta_{\text{reg}}$ and a one-hot vector $\boldsymbol{b}_{\mathrm{\theta}}$ indicating the bin of $\theta$.

\paragraph*{\textbf{Dataset}}
We train our lifting model with the KITTI object detection dataset \cite{6248074}.
As ground truth, we crop the ROIs according to the ground truth of the 2D detections.
The semantic and spatial information is obtained by the methods mentioned in Section \ref{sec:system_overview}.

\begin{figure}[h]
\centering
\fontsize{9}{10}\selectfont
\def\svgwidth{1.\columnwidth}
\begingroup%
  \makeatletter%
  \providecommand\color[2][]{%
    \errmessage{(Inkscape) Color is used for the text in Inkscape, but the package 'color.sty' is not loaded}%
    \renewcommand\color[2][]{}%
  }%
  \providecommand\transparent[1]{%
    \errmessage{(Inkscape) Transparency is used (non-zero) for the text in Inkscape, but the package 'transparent.sty' is not loaded}%
    \renewcommand\transparent[1]{}%
  }%
  \providecommand\rotatebox[2]{#2}%
  \newcommand*\fsize{\dimexpr\f@size pt\relax}%
  \newcommand*\lineheight[1]{\fontsize{\fsize}{#1\fsize}\selectfont}%
  \ifx\svgwidth\undefined%
    \setlength{\unitlength}{169.6996516bp}%
    \ifx\svgscale\undefined%
      \relax%
    \else%
      \setlength{\unitlength}{\unitlength * \real{\svgscale}}%
    \fi%
  \else%
    \setlength{\unitlength}{\svgwidth}%
  \fi%
  \global\let\svgwidth\undefined%
  \global\let\svgscale\undefined%
  \makeatother%
  \begin{picture}(1,0.62541808)%
    \lineheight{1}%
    \setlength\tabcolsep{0pt}%
    \put(0,0){\includegraphics[width=\unitlength,page=1]{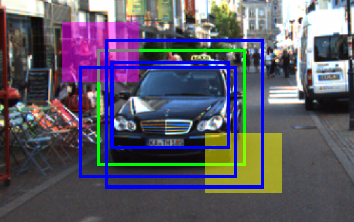}}%
    \put(0.2233613,0.48918732){\color[rgb]{0,0,0}\makebox(0,0)[lt]{\lineheight{1.25}\smash{\begin{tabular}[t]{l}$\boldsymbol{q}_1$\end{tabular}}}}%
    \put(0.67360264,0.11732779){\color[rgb]{0.00392157,0.00392157,0.00392157}\makebox(0,0)[lt]{\lineheight{1.25}\smash{\begin{tabular}[t]{l}$\boldsymbol{q}_2$\end{tabular}}}}%
    \put(0.70380607,0.31021331){\color[rgb]{1,1,1}\makebox(0,0)[lt]{\lineheight{1.25}\smash{\begin{tabular}[t]{l}h\end{tabular}}}}%
    \put(0.47366832,0.17023937){\color[rgb]{1,1,1}\makebox(0,0)[lt]{\lineheight{1.25}\smash{\begin{tabular}[t]{l}l\end{tabular}}}}%
  \end{picture}%
\endgroup%
 \caption{Data augmentation for the lifting process.
The \textcolor{green}{green box} shows the groundtruth 2D bounding box of the vehicle.
The left top and bottom right points are marked as $\boldsymbol{q}_1$ and $\boldsymbol{q}_2$.
Depending on the length $l$ and the width $w$, the possible locations for $q_1^*$ are marked as the \textcolor{darkraspberry}{purple area} and for $q_2^*$ as the \textcolor{gold(web)(golden)}{yellow area}.
Three \textcolor{blue}{blue boxes} are given as examples of possible shifted 2D bounding boxes.
}
\label{fig:dataAug}
\end{figure}
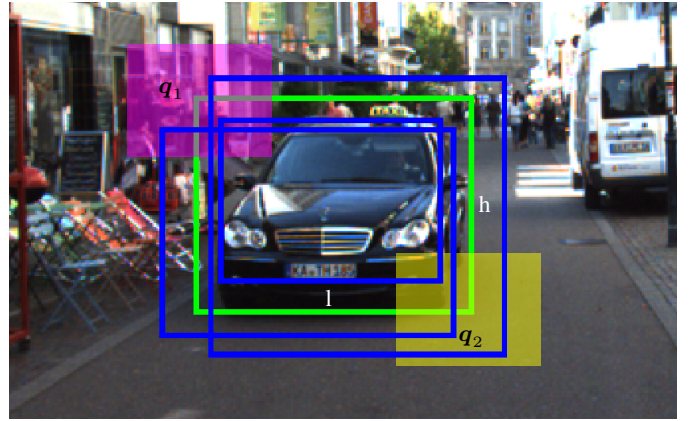

\paragraph*{\textbf{Data Augmentation}} In order to generalize our model performance and avoid overfitting, we conduct data augmentation for our training.
Instead of using ground truth 2D bounding boxes, we vary the shape of the 2D bounding boxes (Figure~\ref{fig:dataAug}).
A 2D bounding box can be determined by a top-left point $\boldsymbol{q}_1$ and a bottom-right point $\boldsymbol{q}_2 = \left [ u_2, v_2 \right ]^{\text{T}}$.
Given a $\boldsymbol{q}_1 = \left [ u_1, v_1 \right ]^{\text{T}}$ and a $\boldsymbol{q}_2 = \left [ u_2, v_2 \right ]^{\text{T}}$ of a grundtruth 2D bounding box, we calculate the height $h$ and the width $w$ of the box, with $h = v_2 - v_1$ and $w=u_2-u_1$.
Then we randomly sample $\Delta h_1, \Delta h_2 \in \left [ -0.25h, 0.25h \right ]$ and a $\Delta w_1, \Delta w_2 \in \left [ -0.25w, 0.25w \right ]$ for the shifting and obtain a new 2D bounding box determined through the $q_1^*$ and $q_2^*$, with $q_1^* = [u_1+\Delta w_1, v_1+\Delta h_1]$ and $q_2^* = [u_2+\Delta w_2, v_2+\Delta h_2]$.

\paragraph*{\textbf{Training with Multi-Task Losses}}
For the regression of $\Delta \boldsymbol{p}$, $\Delta \boldsymbol{d}$ and $\theta_{\text{reg}}$ we use smooth-$L_1$ (huber) loss and obtain $L_{\text{reg},\Delta \boldsymbol{p}}$, $L_{\text{reg},\Delta \boldsymbol{d}}$ and $L_{\text{reg},\theta_{\text{reg}}}$.
For the classification $\boldsymbol{b}_{\theta}$ we use the softmax and cross-entropy and obtain $L_{\text{cls},\boldsymbol{b}_{\theta}}$.
The total loss $L_{\text{total}}$ is defined as:
\begin{equation}
  L_{\text{total}} = \lambda _1L_{\text{reg},\Delta \boldsymbol{p}} + \lambda _2L_{\text{reg},\Delta \boldsymbol{d}} + \lambda _3L_{\text{reg},\theta_{\text{reg}}}+ \lambda _4L_{\text{cls},\boldsymbol{b}_{\theta}}.
\label{eq:loss}
\end{equation}

\begin{figure}[h]
\centering
\fontsize{7}{8}\selectfont
\def\svgwidth{1.25\columnwidth}
\begingroup%
  \makeatletter%
  \providecommand\color[2][]{%
    \errmessage{(Inkscape) Color is used for the text in Inkscape, but the package 'color.sty' is not loaded}%
    \renewcommand\color[2][]{}%
  }%
  \providecommand\transparent[1]{%
    \errmessage{(Inkscape) Transparency is used (non-zero) for the text in Inkscape, but the package 'transparent.sty' is not loaded}%
    \renewcommand\transparent[1]{}%
  }%
  \providecommand\rotatebox[2]{#2}%
  \newcommand*\fsize{\dimexpr\f@size pt\relax}%
  \newcommand*\lineheight[1]{\fontsize{\fsize}{#1\fsize}\selectfont}%
  \ifx\svgwidth\undefined%
    \setlength{\unitlength}{318.81889231bp}%
    \ifx\svgscale\undefined%
      \relax%
    \else%
      \setlength{\unitlength}{\unitlength * \real{\svgscale}}%
    \fi%
  \else%
    \setlength{\unitlength}{\svgwidth}%
  \fi%
  \global\let\svgwidth\undefined%
  \global\let\svgscale\undefined%
  \makeatother%
  \begin{picture}(1,0.34019083)%
    \lineheight{1}%
    \setlength\tabcolsep{0pt}%
    \put(0,0){\includegraphics[width=\unitlength,page=1]{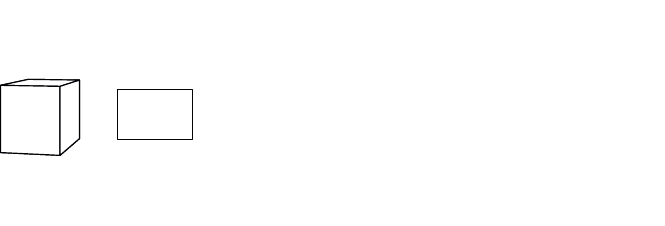}}%
    \put(0.19041597,0.1750277){\color[rgb]{0,0,0}\makebox(0,0)[lt]{\lineheight{1.25}\smash{\begin{tabular}[t]{l}Modified\\ResNet50\end{tabular}}}}%
    \put(0,0){\includegraphics[width=\unitlength,page=2]{netstrcuture_final.pdf}}%
    \put(0.00940454,0.17164831){\color[rgb]{0,0,0}\makebox(0,0)[lt]{\lineheight{1.25}\smash{\begin{tabular}[t]{l}Resized \\Crops\end{tabular}}}}%
    \put(0,0){\includegraphics[width=\unitlength,page=3]{netstrcuture_final.pdf}}%
    \put(0.3452466,0.14941639){\color[rgb]{0,0,0}\rotatebox{-45.38976149}{\makebox(0,0)[lt]{\lineheight{1.25}\smash{\begin{tabular}[t]{l}concatenate\end{tabular}}}}}%
    \put(0,0){\includegraphics[width=\unitlength,page=4]{netstrcuture_final.pdf}}%
    \put(0.07959658,0.29498866){\color[rgb]{0,0,0}\makebox(0,0)[lt]{\lineheight{1.25}\smash{\begin{tabular}[t]{l}$\boldsymbol{p_{\text{m}}}$\end{tabular}}}}%
    \put(0.07959658,0.2371966){\color[rgb]{0,0,0}\makebox(0,0)[lt]{\lineheight{1.25}\smash{\begin{tabular}[t]{l}$\boldsymbol{d}_{\text{prior},k}$\end{tabular}}}}%
    \put(0.04229838,0.00166786){\color[rgb]{0,0,0}\makebox(0,0)[lt]{\lineheight{1.25}\smash{\begin{tabular}[t]{l}One-hot vector for the class\end{tabular}}}}%
    \put(0.55055041,0.31405825){\color[rgb]{0,0,0}\makebox(0,0)[lt]{\lineheight{1.25}\smash{\begin{tabular}[t]{l}MLP\end{tabular}}}}%
    \put(0.68419034,0.25998423){\color[rgb]{0,0,0}\makebox(0,0)[lt]{\lineheight{1.25}\smash{\begin{tabular}[t]{l}$\Delta \textbf{p}$\end{tabular}}}}%
    \put(0.68419034,0.15418033){\color[rgb]{0,0,0}\makebox(0,0)[lt]{\lineheight{1.25}\smash{\begin{tabular}[t]{l}$\Delta \boldsymbol{d}$\end{tabular}}}}%
    \put(0.68419034,0.05048383){\color[rgb]{0,0,0}\makebox(0,0)[lt]{\lineheight{1.25}\smash{\begin{tabular}[t]{l}$\boldsymbol{b}_{\theta}$ + $\theta_{\text{reg}}$\end{tabular}}}}%
  \end{picture}%
\endgroup%
 \caption{Network architecture of the lifting model. 
The network expects four inputs: An one-hot vector indicating the class of detected object, $\boldsymbol{p_{\text{m}}}$, $\boldsymbol{d}_{\text{prior,k}}$ and the resized crops with semantic and spatial information, which are firstly processed by the modified ResNet50. The extracted feature maps are flattened and concatenated with $\boldsymbol{p_{\text{m}}}$, $\boldsymbol{d}_{\text{prior,k}}$ and the one-hot vector. The concatenated feature vectors are passed to three branches (MLPs) to estimate the $\Delta \textbf{p}$, $\Delta \boldsymbol{d}$, $\theta_{\text{reg}}$ and $\boldsymbol{b}_{\theta}$ for the 3D bounding boxes.
}
\label{fig:network}
\end{figure}

Due to different scales of the semantic information ($[0, 1]$) and the spatial coordinates ($[-50\text{m}, 50\text{m}]$), we modify the ResNet50 by removing all the batch normalization layers.
Another option would be to normalize the spatial coordinates to $[0, 1]$ as well.
Since the points would shrink in a narrow space, many geometric features are numerically no longer distinguishable and thereby the geometric features may vanish. 
Additionally, our experiments have proven that the training with normalized coordinates showed poor performance regarding the convergence.
The network architecture is shown in Figure~\ref{fig:network}.

\section{Results and Evaluation}
\label{sec:results_and_evaluation}
\subsection{3D Bounding Box Evaluation}

\begin{table*}[!t]
\centering
\resizebox{.95\textwidth}{!}{
\begin{tabular}{|c|c|c|c|c|c|}
\hline
\multicolumn{2}{|c|}{Metrics}                                                                            & Mono3D\_PLiDAR \cite{weng2019monocular} & ROI-10D \cite{Manhardt_2019_CVPR} & \textbf{Ours with Mono} & \textbf{Ours with Stereo} \\ \hline
\multirow{3}{*}{\begin{tabular}[c]{@{}c@{}}$\text{AP}_{\text{BEV}}$ (in \%)\\ with\\ \textbf{IoU= 0.7}\end{tabular}} & Easy     & 21.27        & 9.78 & 6.93 & 19.63          \\ \cline{2-6}
                                                                                              & Moderate & 13.92        & 4.91 & 4.57         & 12.04          \\ \cline{2-6}
                                                                                              & Hard     & 11.25       & 3.74 & 3.44         & 9.54          \\ \hline
\multirow{3}{*}{\begin{tabular}[c]{@{}c@{}}$\text{AP}_{\text{3D}}$ (in \%)\\ with\\ \textbf{IoU= 0.7}\end{tabular}}  & Easy     & 10.76        & 4.32  & 3.18 & 9.23          \\ \cline{2-6}
                                                                                              & Moderate & 7.50        & 2.02  & 1.78        & 5.68          \\ \cline{2-6}
                                                                                              & Hard     & 6.10        & 1.46  & 1.33        & 4.24          \\ \hline
\end{tabular}
}
\caption{Quantitative comparison on the KITTI test set for car detections.}
\label{tab:3DBBoxCar}
\end{table*}

\begin{table*}[!h]
\centering
\resizebox{.95\textwidth}{!}{
\begin{tabular}{|c|c|c|c|c|c|}
\hline
\multicolumn{2}{|c|}{Metrics}                       & \multicolumn{2}{c|}{$\text{AP}_{\text{BEV}}$ (in \%)                   with \textbf{IoU=0.7} } & \multicolumn{2}{c|}{$\text{AP}_{\text{3D}}$ (in \%) with \textbf{IoU=0.7}}                                       \\ \hline
Class                       & Methods               & RT3DStereo \cite{8917330}            & \textbf{Ours with Stereo}      & RT3DStereo \cite{8917330}            & \textbf{Ours with Stereo}      \\ \hline
\multirow{4}{*}{Pedestrian} & Easy                  & 4.72                  & \textbf{10.84}                 & 3.28                  & \textbf{7.97}                  \\ \cline{2-6}
                            & Moderate              & 3.65                  & \textbf{8.13}                  & 2.45                  & \textbf{5.79}                  \\ \cline{2-6}
                            &Hard & 3.00 &\textbf{6.81} &2.35 &\textbf{4.69} \\ \hline
\multirow{3}{*}{Cyclist}    & Easy             &\textbf{7.03}                  & 4.86                  & \textbf{5.29}                  & 4.65                  \\ \cline{2-6}
                            & Moderate              & \textbf{4.10         }         & 2.78                  & \textbf{3.37}                  & 2.62                  \\ \cline{2-6}
                            & Hard                  & \textbf{3.88                 } & 2.82                  & \textbf{2.57}                  & 2.50                  \\ \hline
\end{tabular}
}
\caption{Quantitative comparison on the KITTI test set for pedestrian and cyclist detections.}
\label{tab:3DBBoxPedandCyc}
\end{table*}

We evaluate our approach on the KITTI object detection
benchmark \cite{6248074}, which has 7481 training images with available ground truth labels and 7518 testing images.
We generate point clouds using the VNLNet \cite{Yin_2019_ICCV} for mono vision and the PSMNet \cite{8578665} for stereo vision and utilize them further for our lifting process. During training, we set the hyperparameters in the loss function (Equation \ref{eq:loss}) to $\lambda _1=\lambda _2=\lambda _3=\lambda _4=1$.
We compare our results with Mono3D\_PLiDAR \cite{weng2019monocular} and ROI-10D \cite{Manhardt_2019_CVPR} on the KITTI test dataset for car detection (Table~\ref{tab:3DBBoxCar}).

Despite the worse 2D detections we use for runtime reasons and the fact that they fine-tuned their instance segmentation on the KITTI instance segmentation benchmark, we achieve results in the range of Mono3D\_PLiDAR \cite{weng2019monocular} with our stereo approach.
In addition, our lifting model with stereo vision outperforms the ROI-10D \cite{Manhardt_2019_CVPR} for all difficulties.
Regarding the moderate and hard metric, the average precision of the bird's eye view $\text{AP}_{\text{BEV}}$ and of the 3D detection $\text{AP}_{\text{3D}}$ is almost tripled, compared to ROI-10D.
Further we recognize, that our results with stereo vision are almost three times better than our results with mono vision, even though VNL \cite{Yin_2019_ICCV} has a higher ranking for depth estimation on KITTI than PSMNet \cite{8578665}.

Since there is no data for pedestrian and cyclist detection of the above-mentioned approaches on KITTI, we compare our pedestrian and cyclist detection with RT3DStereo \cite{8917330}, which also uses stereo vision for object detection (Table~\ref{tab:3DBBoxPedandCyc}).
It is noticeable that our lifting model outperforms RT3DStereo on pedestrian detection.
For all difficulties of pedestrian detection, our $\text{AP}_{\text{BEV}}$ and $\text{AP}_{\text{3D}}$ reach about twice the values of RT3DStereo.
Our performance for the 3D cyclist detection is in the range of RT3DStereo.

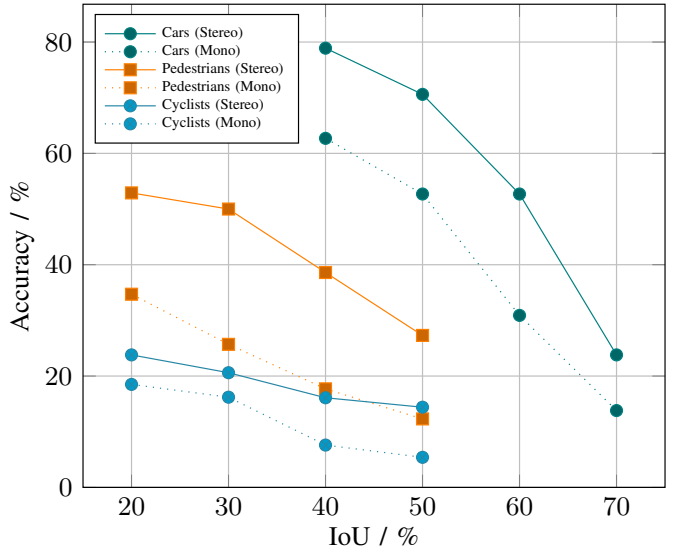
\begin{figure}[h]
\centering
\hspace*{-0.35cm}
\resizebox{1.03\columnwidth}{!} {
\begin{tikzpicture}
\begin{axis}[
	xlabel=IoU / \%,
	ylabel=Accuracy / \%,
	ytick={0,20,...,100},
	ymin=0,
	grid=both,
	x label style={at={(axis description cs:.5 ,-.05)}, anchor=south, font=\small},
	y label style={at={(axis description cs:0.075 ,0.3)}, anchor=west, font=\small},
    tick label style={font=\small},
    cycle multi list={
    exotic\nextlist
    solid,{dotted,mark options={solid}}
    },
    legend cell align={left},
    legend style={at={(0.02,0.98)}, anchor=north west, font=\tiny, row sep=-0.1cm},
]

\addplot plot coordinates {(40, 78.9)(50, 70.6)(60, 52.7)(70, 23.8)};
\addplot plot coordinates {(40, 62.7)(50, 52.7)(60, 30.9)(70, 13.8)};
\addplot plot coordinates {(20, 52.9)(30, 50.0)(40, 38.6)(50, 27.3)};
\addplot plot coordinates {(20, 34.7)(30, 25.7)(40, 17.7)(50, 12.3)};
\addplot plot coordinates {(20, 23.8)(30, 20.6)(40, 16.1)(50, 14.4)};
\addplot plot coordinates {(20, 18.5)(30, 16.2)(40, 7.6)(50, 5.4)};

\legend{Cars (Stereo)\\Cars (Mono)\\Pedestrians (Stereo)\\Pedestrians (Mono)\\Cyclists (Stereo)\\Cyclists (Mono)\\}
\end{axis}
\end{tikzpicture}
}
\caption{Overall validation accuracy depending on the Intersection-over-Union (IoU) in the easy benchmark.}
\label{fig:iou_acc}
\end{figure}

For the reasons addressed in \cite{weng2019monocular} and \cite{8917330}, an Intersection over Union (IoU) of 0.7 for cars is difficult to achieve with camera-based approaches.
Therefore, we study the relationship between the IoU and the detection accuracy ($\text{AP}_{\text{3D}}$).
We obtain the $\text{AP}_{\text{3D}}$ for different IoUs from our validation set for the easy difficulty.
The results are shown in Figure \ref{fig:iou_acc}.
A high $\text{AP}_{\text{3D}}$ ($ >70\%$) can be achieved for the car detection using stereo with $\text{IoU} = 0.5$. This corresponds to a longitudinal and lateral position deviation in the range of only about $\SI{0.5}{\meter}$, which is sufficient for high-level scene understanding and trajectory planning for automated driving.
For pedestrians the $\text{AP}_{\text{3D}}$ can reach almost $50\%$ for an $\text{IoU}=0.3$. For cyclists our average precision is around $20\%$ for all IoUs. Note, that most other approaches ignore pedestrians and cyclists at all, despite they are the most vulnerable road users.

\subsection{Orientation Estimation}
Besides the bounding boxes, we also evaluate the performance of the orientation estimation. For a proper comparison of all three classes, we choose RT3DStereo \cite{8917330} as our baseline.
The results for orientation estimation are provided in terms of average orientation similarity (AOS) and shown in Table~\ref{tab:orientation}.
In general, the results of lifting with mono and stereo are very close to each other and outperform the RT3DStereo \cite{8917330} for all classes and difficulties, especially for car and cyclist.
A higher accuracy of orientation estimation makes predictions of other traffic participants more plausible and increases safety.

\begin{figure}[t]
	\begin{subfigure}{\columnwidth}
		\setlength{\abovecaptionskip}{1pt plus 2pt minus 2pt}
		\setlength{\belowcaptionskip}{4pt plus 2pt minus 2pt}
		\includegraphics[width=\linewidth]{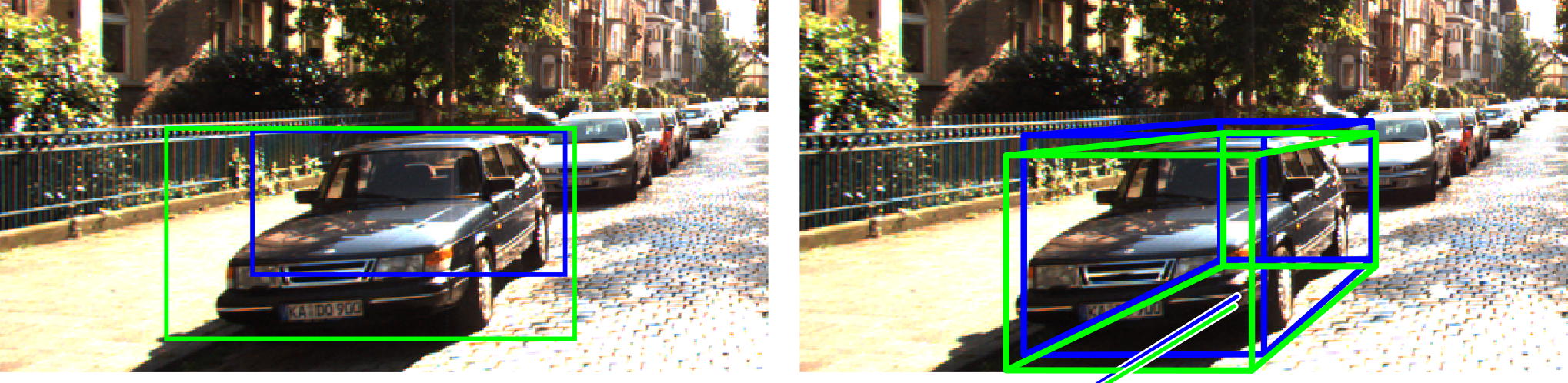}
		\caption{}
		\label{fig:shiftInvariance_a}
	\end{subfigure}
	\begin{subfigure}{\columnwidth}
		\setlength{\abovecaptionskip}{3pt plus 2pt minus 2pt}
		\setlength{\belowcaptionskip}{4pt plus 2pt minus 2pt}
		\includegraphics[width=\linewidth]{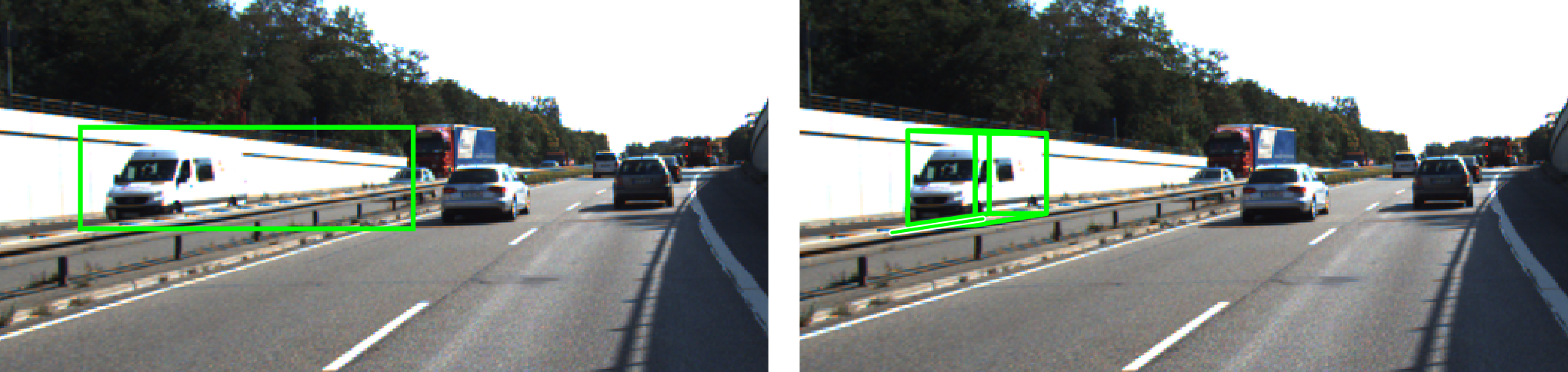}
		\caption{}
		\label{fig:shiftInvariance_b}
	\end{subfigure}
	\begin{subfigure}{\columnwidth}
		\setlength{\abovecaptionskip}{-13pt plus 2pt minus 2pt}
		\setlength{\belowcaptionskip}{8pt plus 2pt minus 2pt}
		\includegraphics[width=\linewidth]{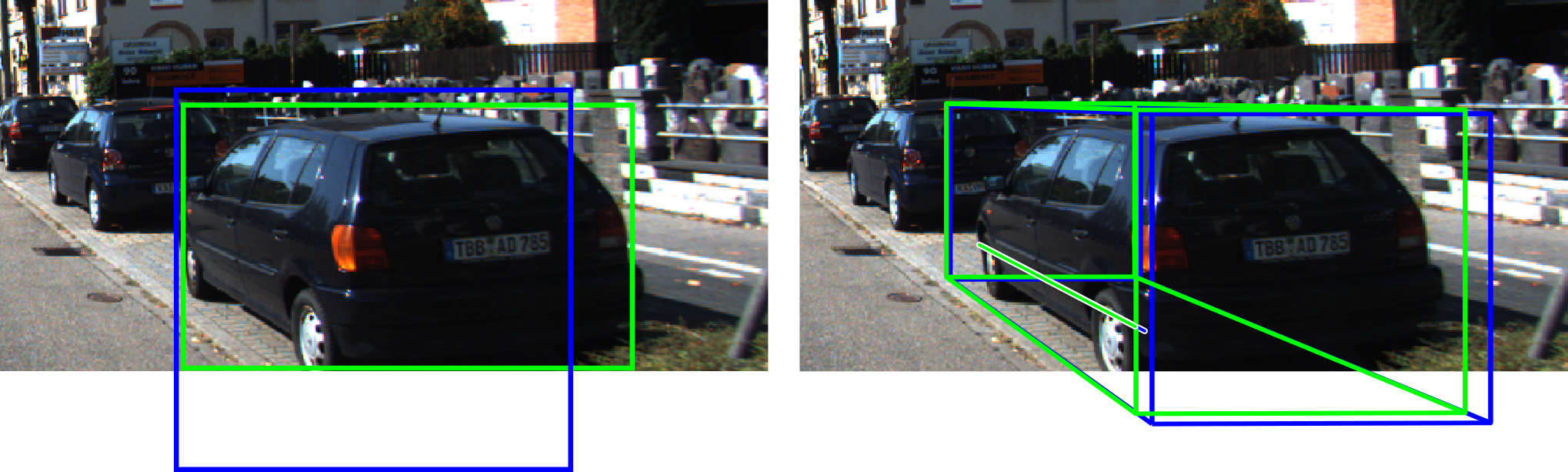}
		\caption{}
		\label{fig:shiftInvariance_c}
	\end{subfigure}
	\begin{subfigure}{\columnwidth}
		\setlength{\abovecaptionskip}{-10pt plus 2pt minus 2pt}
		\setlength{\belowcaptionskip}{0pt plus 2pt minus 2pt}
		\includegraphics[width=\linewidth]{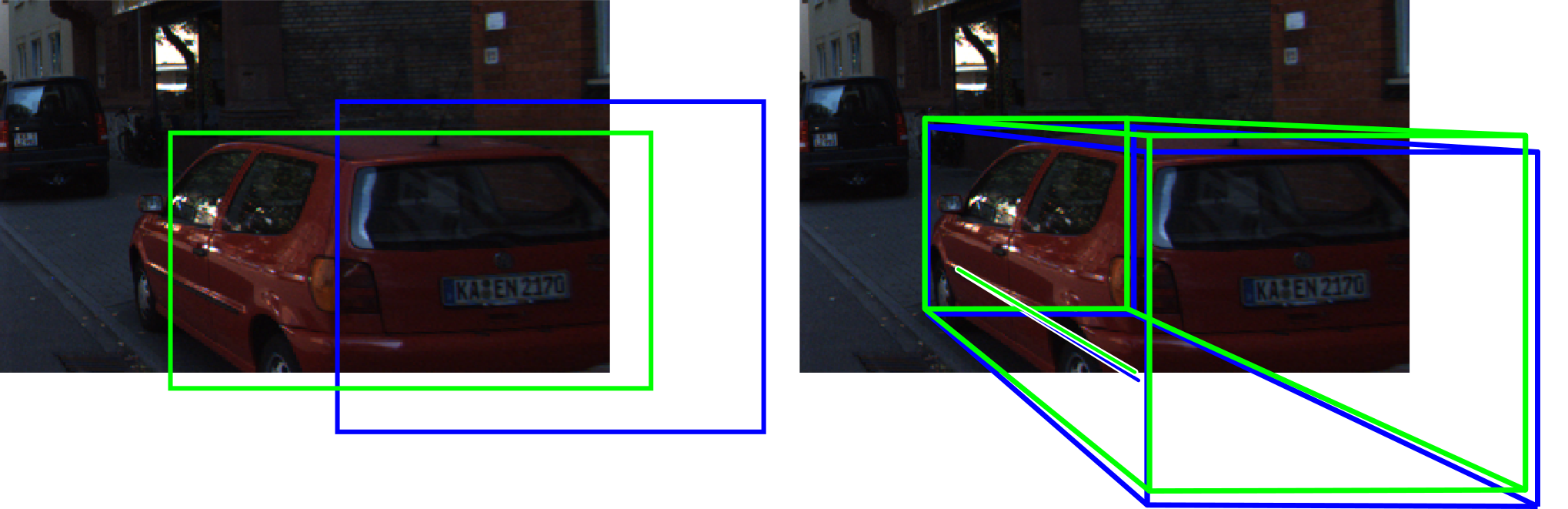}
		\caption{}
		\label{fig:shiftInvariance_d}
	\end{subfigure}
	\caption{Qualitative evaluation of the shift invariance. The left images show different 2D bounding boxes for one object (\textcolor{green}{green} and \textcolor{blue}{blue}), in the right you can see the corresponding 3D bounding boxes with there orientations.}
	\label{fig:shiftInvariance}
\end{figure}

\begin{table}[!t]
\large
\resizebox{\columnwidth}{!}{
\begin{tabular}{|c|c|c|c|c|}
\hline
\multicolumn{2}{|c|}{\multirow{2}{*}{Class}} & \multicolumn{3}{c|}{AOS (in \%)}                                                                                                   \\ \cline{3-5}
\multicolumn{2}{|c|}{}                       & RT3DStereo \cite{8917330} & \begin{tabular}[c]{@{}c@{}} \textbf{Ours with}\\ \textbf{Mono}\end{tabular} & \begin{tabular}[c]{@{}c@{}} \textbf{Ours with}\\ \textbf{Stereo}\end{tabular} \\ \hline
\multirow{3}{*}{Car}           & Easy        & 25.58      & 66.84                                                    & \textbf{66.92}                                                      \\ \cline{2-5}
                               & Moderate    & 21.41      & \textbf{46.13}                                                    & 45.84                                                      \\ \cline{2-5}
                               & Hard        & 17.52      & \textbf{36.73}                                                    & 36.43                                                      \\ \hline
\multirow{3}{*}{Pedestrian}    & Easy        & 21.41      & 21.50                                                    & \textbf{22.15}                                                      \\ \cline{2-5}
                               & Moderate    & 15.34      & 15.60                                                    & \textbf{15.84}                                                      \\ \cline{2-5}
                               & Hard        & 13.23      & 14.18                                                    & \textbf{14.34}                                                      \\ \hline
\multirow{3}{*}{Cyclist}       & Easy        & 5.46       & 13.31                                                    & \textbf{14.49}                                                      \\ \cline{2-5}
                               & Moderate    & 3.88       & \textbf{8.81}                                                     & 8.56                                                       \\ \cline{2-5}
                               & Hard        & 3.54       & 7.43                                                     & \textbf{8.18}                                                       \\ \hline
\end{tabular}
}
\caption{Quantitative comparison on the KITTI test set for orientation estimation.}
\label{tab:orientation}
\end{table}

\subsection{Shift Invariance for Lifting}
The accuracy of the 2D bounding boxes is fundamental for our lifting process (Figure~\ref{fig:pipeline}).
They can be slightly shifted in a horizontal or vertical direction, compared to the ground truth.
Therefore, it is crucial to strengthen the shift-invariance of our lifting process and make it less sensitive to the shifting of 2D bounding boxes.
As addressed in Section \ref{sec:lifting_2d_detection_to_3d}, we utilize the data augmentation during training, where we randomly shift the ground truth 2D bounding boxes from the training set and use them as input for our pipeline.

Depending on the threshold of Non-Maximum Suppression, there can be several 2D bounding boxes belonging to the object during inference.
In our qualitative evaluation, we utilize this to analyze the shift-invariance property of our lifting model.
In Figure~\ref{fig:shiftInvariance_a}, we observe that the lifted 3D bounding boxes of redundant 2D bounding boxes still share a high IoU.
Their estimated orientations, locations and dimensions deviate only minimal from each other.
Figure~\ref{fig:shiftInvariance_b} shows that our model even provides promising 3D bounding boxes for inaccurate 2D bounding boxes.
From Figure~\ref{fig:shiftInvariance_c} and \subref{fig:shiftInvariance_d}, we infer that shift-invariance also performs well for truncated objects.
For the objects that are partially out of boundary, it is impossible to obtain a perfect 2D bounding box or instance mask.
Thus, the Bounding Box Consistency Optimization proposed in \cite{weng2019monocular} cannot be considered in these situations.
Nevertheless, our network can still estimate promising 3D bounding boxes directly for these truncated objects.
We conclude that our data augmentation method improves the shift-invariance and the generalization ability of our lifting network.

\vspace{0.27cm} %
\section{Conclusion and Future Work}
\label{sec:conclusion}
In this paper, we proposed a method based only on stereo or mono vision to lift 2D detections to 3D by utilizing the semantic and spatial information of 2D ROI.
Since the generated point clouds can be organized in an image format, we are the first proposing an approach to process the point clouds with semantic information using a 2D CNN and achieved comparable results for car detection with a related approach \cite{weng2019monocular}.
In our evaluation on the KITTI object detection benchmark we showed that our approach outperforms current state-of-the-art for vision-based pedestrian detection. Good results are also achieved for cyclists.
We have shown that through data augmentation during the training, our lifting process is shift-invariant, indicating a stable performance during the inference. Our evaluation also shows, that lifting with stereo vision is superior to lifting with mono vision.

Since ResNet was designed for images with three channels, we plan to extend the ResNet structure to input formats with $C+3$ channels in order to improve the lifting performance. For flexibility reasons, we have chosen to perform semantic segmentation for each individual ROI. However, some 2D object detectors like \cite{salscheider2019simultaneous} simultaneously provide pixel-wise semantic information for the whole image. This can further reduce the runtime of our entire pipeline. 
\balance
\printbibliography

\end{document}